\documentclass[10pt]{article} 
\usepackage[accepted]{rlc}

\usepackage{amssymb}            
\usepackage{mathtools}          
\usepackage{mathrsfs}           
\mathtoolsset{showonlyrefs}     
\usepackage{graphicx}           
\usepackage{subfigure}
\usepackage{subcaption}         
\usepackage[space]{grffile}     
\usepackage{url}    
\usepackage{authblk}

\title{Sophia: A Persistent Agent Framework of Artificial Life}

\author[1,2]{\textbf{Mingyang Sun}}
\author[3]{\textbf{Feng Hong}}
\author[2,4*]{\textbf{Weinan Zhang}}

\affil[]{Westlake University, \textsuperscript{2}Shanghai Innovation Institute}
\affil[3]{Project Cuddlepark Team, 
\textsuperscript{4}Shanghai Jiao Tong University
}

\affil[ ]{\texttt{sunmingyang@westlake.edu.cn, wnzhang@sjtu.edu.cn}}

\begin{document}
\begin{center}
\maketitle
\end{center}
\begin{abstract}
The rapid development of Large Language Models (LLMs) has elevated AI agents from task-specific tools to long-lived, decision-making entities capable of independent planning and strategic collaboration. However, most existing architectures remain reactive: they rely on manually crafted configurations that remain static after deployment, designed for narrow tasks or fixed scenarios. These systems excel at perception (System 1) and deliberation (System 2), yet lack a persistent meta-layer capable of maintaining identity, verifying internal reasoning, and aligning short-term tasks with long-term survival.

In this work, we first propose a third stratum, System 3, that presides over the agent's narrative identity and long-horizon adaptation. The framework maps selected psychological constructs (e.g., meta-cognition, theory-of-mind, intrinsic motivation, episodic memory) to concrete computational modules, thereby translating abstract notions of artificial life into implementable design requirements. These ideas coalesce in Sophia, a ``Persistent Agent'' wrapper that grafts a continuous self-improvement loop onto any LLM-centric System 1/2 stack. Sophia is driven by four synergistic mechanisms: process supervised thought search that curates and audits emerging thoughts, an memory module that maintains narrative identity, dynamic user and self models that track external and internal beliefs,  and a hybrid reward system balancing environmental feedback with introspective drives. Together, transform the highly repetitive reasoning episodes of a primitive agent into an endless, self-driven reasoning phase dedicated to diverse goals, enabling autobiographical memory,  identity continuity, and transparent narrative explanations of behavior.

Although the paper is primarily conceptual, grounding System 3 in decades of cognitive theory, we provide a compact engineering prototype to anchor the discussion. In a deployment spanning prolonged durations within a dynamic web environment, Sophia demonstrated robust operational persistence through autonomous goal generation. Quantitatively, Sophia independently initiates and executes various intrinsic tasks while achieving an 80\% reduction in reasoning steps for recurring operations. Notably, meta-cognitive persistence yielded a 40\% gain in success for high-complexity tasks, effectively bridging the performance gap between simple and sophisticated goals. Qualitatively, System 3 exhibited a coherent narrative identity and an innate capacity for task organization.  By fusing psychological insight with a lightweight reinforcement-learning core, the persistent agent architecture advances a possible practical pathway toward artificial life.

\noindent{\quad}

\textbf{Keywords:} Persistent Agent, LLM Agents, System 3, Artificial Life 
\end{abstract}

\section{Introduction}
\label{sec:introduction}
The rapid proliferation of large language models (LLMs) has catalyzed a paradigm shift in AI agents, transforming them from single task executors into long-lived sophisticated cognitive entities endowed with capabilities for autonomous planning, strategic deliberation, and collaborative engagement~\citep{achiam2023gpt, grattafiori2024llama, hurst2024gpt, guo2025deepseek, yang2025qwen3}. This technological leap is reshaping expectations across science, industry, and everyday applications~\citep{yang2024finrobot, chkirbene2024large, ren2025towards, li2024personal}. Yet despite these breakthroughs, most existing agent frameworks remain anchored to manually crafted configurations that remain static after deployment. Once shipped, they cannot revise their skill set, develop new tasks, or integrate unfamiliar knowledge without human engineers in the loop. Lacking the intrinsic motivation and self-improvement capabilities inherent to living systems, today's agents remain unable to achieve sustained growth or open-ended adaptation. Infusing AI agents with these lifelike principles, enabling autonomous self-reconfiguration while maintaining operational coherence, has thus emerged as a critical frontier in AI research.

Within prevailing agent architectures, cognition is typically partitioned into two complementary subsystems~\citep{li2025system}. System 1 embodies rapid, heuristic faculties—perception, retrieval, and instinctive response. System 2, by contrast, governs slow, deliberate reasoning. It employs chain-of-thought planning, multi-step search, counterfactual simulation, and consistency checks to refine or override System 1’s impulses. In practical LLM agents, this often manifests as a reasoning loop that expands prompts with scratch-pad deliberations, validates tool outputs, and aligns final responses with user goals. While the synergy of these two layers enables impressive task performance, both remain confined to static configurations and predetermined task scheduling. Even in cases where agents support continual learning, such updates typically follow an externally defined task schedule rather than being self-directed. Consequently, the agent can neither update its reflexive priors nor revise its thought process when encountering truly novel domains.

This rigidity highlights the necessity of a higher-order "\textbf{System 3}", which is a meta-cognitive layer that monitors, audits, and continuously adapts both underlying systems, thereby enabling the entire cognitive architecture to sustain ongoing learning. In this work, we ground System 3 in four foundational theories from cognitive psychology:
\begin{itemize}
    \item Meta-cognition~\citep{shaughnessy2008meta, dunlosky2008metacognition}: a self-reflective monitor that inspects ongoing thought traces, flags logical fallacies, and selectively rewrites its own procedures;
    \item Theory-of-Mind~\citep{frith2005theory, wellman2018theory}: an explicit model of actors (humans or agents) that infers their beliefs, desires, and intentions to guide cooperation and learning;
    \item Intrinsic Motivation~\citep{fishbach2022structure}: an internal reward generator that balances extrinsic task success with curiosity-driven exploration, enabling the agent to prioritize long-term competence over short-term gains.
    \item Episodic Memory~\citep{tulving2002episodic, ezzyat2011constitutes}: A structured autobiographical record that stores, indexes, and retrieves past experiences, providing crucial context for interpreting current events and planning future actions.
\end{itemize}
By integrating these components into a persistent control loop, System 3 turns an otherwise static agent into a self-aware learner that can not only reason about the world but also reason about—and iteratively improve—its own reasoning process.

Such a layer is indispensable for three fundamental reasonsespecially in the context of artificial life~\citep{langton1997artificial, langton2019artificial}. First, real-world environments are non-stationary: objectives shift, constraints evolve, and previously unseen tasks emerge without warning. Agents devoid of self-reconfiguration inevitably ossify, leading to performance decay or catastrophic failure. Second, sustained autonomy demands identity continuity; without a mechanism to maintain a coherent self-model across sessions, an agent cannot accumulate autobiographical knowledge, assess longitudinal progress, or ensure behavioral consistency. Third, safety and alignment require transparent introspection: only a meta-cognitive agent can audit its decision pathways in real time and correct misaligned incentives before they propagate into harmful actions. In short, System 3 elevates agents from transient problem-solvers to adaptive, trustworthy partners capable of lifelong learning in open-ended environments.

To operationalize System 3 in a deployable setting, we present the persistent agent, Sophia, which is a compact, modular framework that endows any LLM-centric System 1/2 stack with a continual self-improvement loop. The design hinges on the following reinforcing mechanisms. 
\begin{itemize}
    \item Process-Supervised Thought Search captures raw chain-of-thought traces, filters them through self-critique prompts, and stores only validated reasoning paths for future reuse, turning stochastic deliberations into reusable cognitive assets.
    \item A Memory Module maintains a structured memory graph of goals, experiences, and self-assessments, giving the agent a stable narrative identity that persists across reboots and task domains.
    \item A Self-Model records the agent’s capabilities, terminal creed and intrinsic state; gaps detected here are immediately feedback as new learning targets. A User-Model maintains a live belief state for each user—goals, social relationship and human preferences.
    \item A Hybrid Reward Module blends external task feedback with intrinsic signals—curiosity, coherence, and self-consistency—so the agent not only pursues immediate goals but also maximizes long-term competence.
\end{itemize}

Together, these components create an end-to-end meta-cognitive loop: the agent plans, acts, reflects on its performance, updates its procedures, and re-aligns future behavior without human intervention. 

This work makes the following key contributions:
\begin{itemize}
    \item We introduce the conceptualization of a System 3 architecture for AI agents, grounded in integrative cognitive psychological foundations including meta-cognition, theory of mind, intrinsic motivation, and episodic memory.
    \item We present Sophia, the first computationally realizable agent system for artificial life capable of generating its own learning goals, curating personalized skill curricula, and sustaining autonomous self-adaptation without external task scheduling or reward engineering.
    \item We demonstrate, through a 24-hour continuous deployment in a web simulation environment, that our approach enables sustained autonomy, coherent identity persistence, and open-ended competency growth—marking a significant step toward artificial agents that exhibit lifelike learning and self-evolution.
\end{itemize}

\section{Related Work}

\subsection{Continual Learning}
Our work builds upon yet significantly extends the field of Continual Learning (CL)~\citep{wang2024comprehensive}, also known as lifelong learning, which aims to enable machine learning models to learn sequentially from a stream of data while mitigating the problem of catastrophic forgetting—where learning new tasks causes abrupt degradation of performance on previously learned ones. Prominent strategies include architectural approaches (e.g., adding new parameters or modules)~\citep{lu2024revisiting}, regularization-based methods~\citep{ahn2019uncertainty} that constrain weight updates to protect important parameters for old tasks, and memory-replay methods~\citep{chaudhry2019continual, lopez2017gradient} that maintain a small buffer of past examples for rehearsal. 

In the context of large language models (LLMs), continual learning has gained significant attention~\citep{fang2025comprehensive, wang2024comprehensive}. \cite{zheng2025towards} provide a comprehensive survey specifically focused on lifelong learning methods for LLMs, categorizing approaches into internal and external knowledge strategies. \cite{ke2022continual} explore continual learning in natural language processing tasks, emphasizing techniques to prevent catastrophic forgetting and enable knowledge transfer. Recently, continual pre-training, instruction tuning, and alignment strategies for LLMs have attracted more attention from researchers~\citep{zhou2024continual, zheng2025lifelong}.

\begin{figure}[ht]
    \begin{center}

        \includegraphics[trim=10mm 13mm 30mm 120mm, clip, width=0.95
  \linewidth]{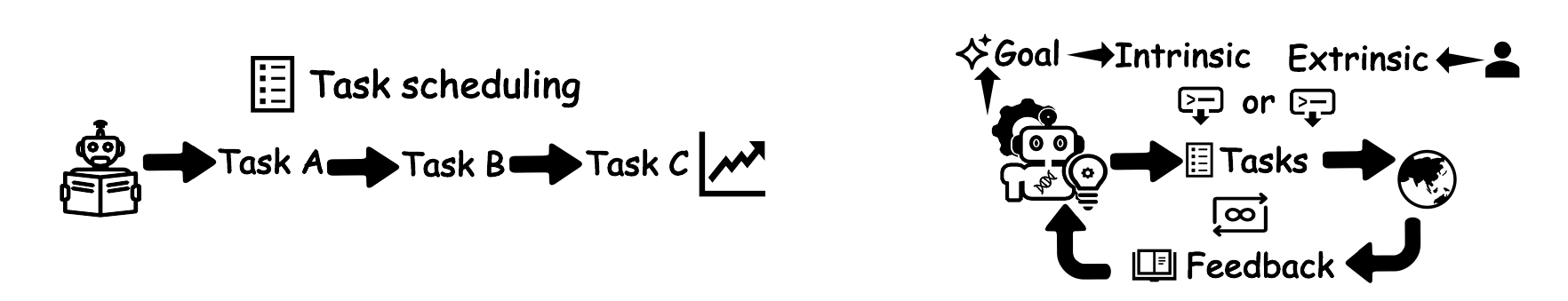}
  \begin{minipage}{0.5\textwidth}
    \centering
    \textbf{\quad(a) Continual Learning of Agent.}
  \end{minipage}\hfill
  \begin{minipage}{0.46\textwidth}
    \centering
    \textbf{(b) Persistent Agent for Artificial Life.} 
    
  \end{minipage}
    \end{center}
    \caption{Paradigm comparison of continual learning and persistent agent. (a) Continual-learning agents follow an externally defined schedule to update their model only when a new assignment is pushed to them.
(b) A persistent agent runs an internal goal-feedback loop: it autonomously selects goals, acts in the environment, evaluates the outcome, and refines its next goals, enabling open-ended, self-directed adaptation.}
    \label{fig:CLvsAL}
\end{figure}

While the existing methods have achieved considerable success in predefined task sequences, they primarily operate within a static learning paradigm where the objectives, task boundaries, and data distributions are externally defined and provided to the model. In contrast, our concept of a persistent agent subsumes continual learning as one component of a larger cognitive architecture, see Figure~\ref{fig:CLvsAL}. Whereas CL systems are typically passive learners that acquire skills from an externally curated task scheduling, a persistent agent is an active, self-directed learner. It not only mitigates forgetting but also autonomously generates its own goals, constructs its own learning curricula, and governs its own learning process through meta-cognitive control. This shift, from merely adapting to a given data stream to proactively seeking knowledge and self-improvement, represents a fundamental evolution from passive continual learning towards autonomous, open-ended development.

\subsection{Forward Learning and Backward Learning in LLMs}
The learning processes in Large Language Models (LLMs) can be conceptually categorized into two paradigms: forward learning and post training. This distinction is crucial for understanding how our persistent agent achieves continuous adaptation.

\textbf{Forward Learning} refers to the model's ability to acquire and internalize new knowledge during the inference stage, without any weight updates. This is primarily achieved through in-context learning~\citep{li2023practical}, where a model conditions on a prompt containing demonstrations or new information and immediately applies this context to complete the task at hand. While powerful for few-shot adaptation, forward learning is transient; the knowledge is ephemeral and confined to the current session, leaving no lasting trace in the model's parameters.

\textbf{Post Training}, in contrast, denotes the traditional process of updating the model's weights based on new data, typically through fine-tuning~\citep{zhang2024scaling, wu2025llm} or reinforcement learning from human feedback (RLHF)~\citep{bai2022training, lambert2025reinforcement}. This form of learning results in persistent, long-term changes to the model's behavior and knowledge base. However, it is often computationally expensive, requires careful curation of datasets to avoid catastrophic forgetting, and typically occurs in offline batches rather than continuously.

Our persistent agent architecture seamlessly integrates both paradigms. In our architecture, the sub-modules of System 3 collectively establish a dynamic contextual foundation throughout the agent’s lifelong operation, enabling rapid, on-the-fly adaptation and reasoning via forward learning. Meanwhile, when capability gaps are identified—such as insufficient reasoning ability or skill deficiencies—System 2 can undergo enhancement through backward learning, guided by the reward model within System 3 that aligns updates with meta-cognitive goals and intrinsic motivations.

\section{The Psychological Pillars of System 3}\label{sec:system3}

The ambition to evolve AI agents from sophisticated tools into digital beings necessitates a leap in cognitive architecture. We posit that this leap requires a System 3—a supervisory cognitive layer responsible for integration, self-reflection, and long-term coherence. Whereas System 1 delivers rapid, heuristic responses and System 2 performs deliberate analytical reasoning, System 3 orchestrates the meta-cognitive processes that underpin a sense of self and enduring purpose.  To ground this architecture in a plausible model of intelligence, we draw on four foundational constructs from cognitive psychology—\textit{Theory of Mind}, \textit{Episodic Memory}, \textit{Meta-Cognition} (with a \textit{Self-Model}) and \textit{Intrinsic Motivation}. These concepts provide the necessary framework for an agent to not only think but to think about its own thinking, to learn from its experiences, and to act with a degree of autonomy that transcends predefined tasks.

The interplay of these core components can be visualized in the following diagram, which outlines the core information flow within our proposed System 3 architecture:

The Meta-Cognitive Monitor—the core of System 3-ingests salient events from the environment, consults four psychological modules, and issues updates or directives to Systems 2 and 1. 

\begin{figure}[ht]
    \begin{center}
        \includegraphics[trim=0mm 3mm 0mm 20mm, clip, width=0.95
  \linewidth]{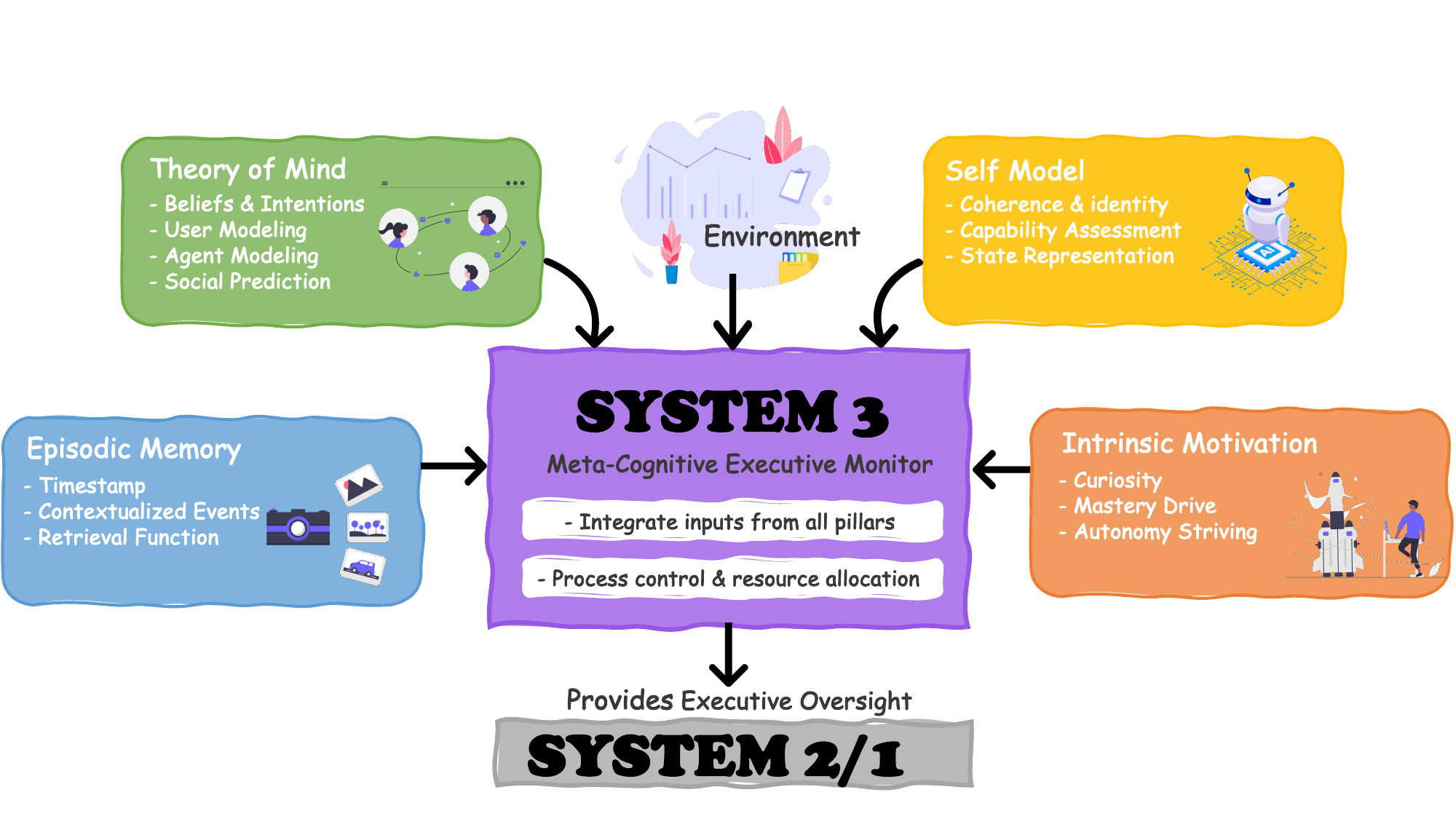}
    \end{center}
    \caption{Schematic of the proposed System 3 architecture. A central Meta-Cognitive Executive Monitor receives salient events from the environment and integrates signals from four psychological pillars—Theory of Mind (belief and intention modeling, user/agent prediction), Episodic Memory (timestamped, context-rich event store with retrieval), Self Model (coherence, capability assessment, state representation), and Intrinsic Motivation (curiosity, mastery drive, autonomy striving). Guided by this aggregated context, the monitor performs process control and resource allocation, issuing executive oversight commands to the underlying System 2/1 perception-and-reasoning stack.}
    \label{fig:system3}
\end{figure}

\subsection{Foundational Constructs}

\textit{Theory of Mind} is the capacity to attribute mental states—such as beliefs, intents, and knowledge—to others. For a persistent agent, this transcends simple user intent recognition. It allows the agent to build rich, dynamic models of the individuals it interacts with, anticipating their needs and reactions. This capability is fundamental for engaging in truly collaborative and adaptive interactions, as the agent can tailor its communication and actions based on its understanding of the user's perspective.

\textit{Episodic Memory} provides the substrate for an autobiographical self. Rather than storing facts in isolation, this system records experiences as contextualized events, complete with temporal and situational markers. This allows the agent to construct a narrative of its own history, which is indispensable for maintaining consistency over time. By recalling past successes and failures in full context, the agent can learn in a more nuanced way, avoiding past errors and building upon strategies that have proven effective. To keep storage tractable, a tiered retrieval scheme is neccesary: high-level summaries for fast search, with raw traces lazily retrieved only when relevance exceeds a threshold.

\textit{Meta-Cognition} with \textit{Self-Model} form the executive heart of System 3. Meta-cognition is the practice of monitoring and regulating one's own cognitive processes. This is enabled by a Self-Model—an internal representation of the agent's own capabilities, performance, and current state. Meta-Cognitive Monitor supervises the entire reasoning pipeline, setting open-ended goals, allocating resources (e.g., deciding when to invoke deep System 2 search), detecting logical fallacies and triggering Self-Model revisions.

Finally, to transition from a reactive system to a proactive one, an agent requires a drive to act beyond explicit instruction. This is achieved through \textit{Intrinsic Motivation}. We implement drives such as Curiosity (a desire to seek novel information and reduce uncertainty), Mastery (a urge to develop competence and solve increasingly complex problems), and Relatedness (a drive to establish and maintain meaningful connections with users). These internal motivators provide the energy and direction for exploratory behavior and long-term goal pursuit, ensuring the agent is not merely a passive tool but an engaged participant in its environment. Conflicts among intrinsic drives and external task rewards are resolved through a hierarchical planner that learns a dynamic weighting scheme, preserving safety and alignment.

\subsection{Why These Pillars Matter}
Real-world environments evolve. Objectives shift, constraints change and novel tasks surface unexpectedly. Without a System 3 layer, static agents ossify, leading to performance decay or catastrophic failure. The four pillars above let an agent:
\begin{itemize}
    \item Adapt: Theory-of-Mind and Intrinsic Motivation guide exploration that is both socially aware and autonomy striving;
    \item Accumulate: Episodic Memory grounds learning in lived history, enabling long-horizon credit assignment.
    \item Audit: Meta-Cognition supplies real-time self-inspection, a prerequisite for safety and transparent alignment.
\end{itemize}

In concert, they transform an LLM-centric System 1/2 stack into a persistent, self-improving entity capable of lifelong learning and trustworthy collaboration.


\section{The Persistent Agent with System 3 in Operation}
\begin{figure}[ht]
    \begin{center}
        \includegraphics[trim=0mm 0mm 0mm 10mm, clip, width=0.98
  \linewidth]{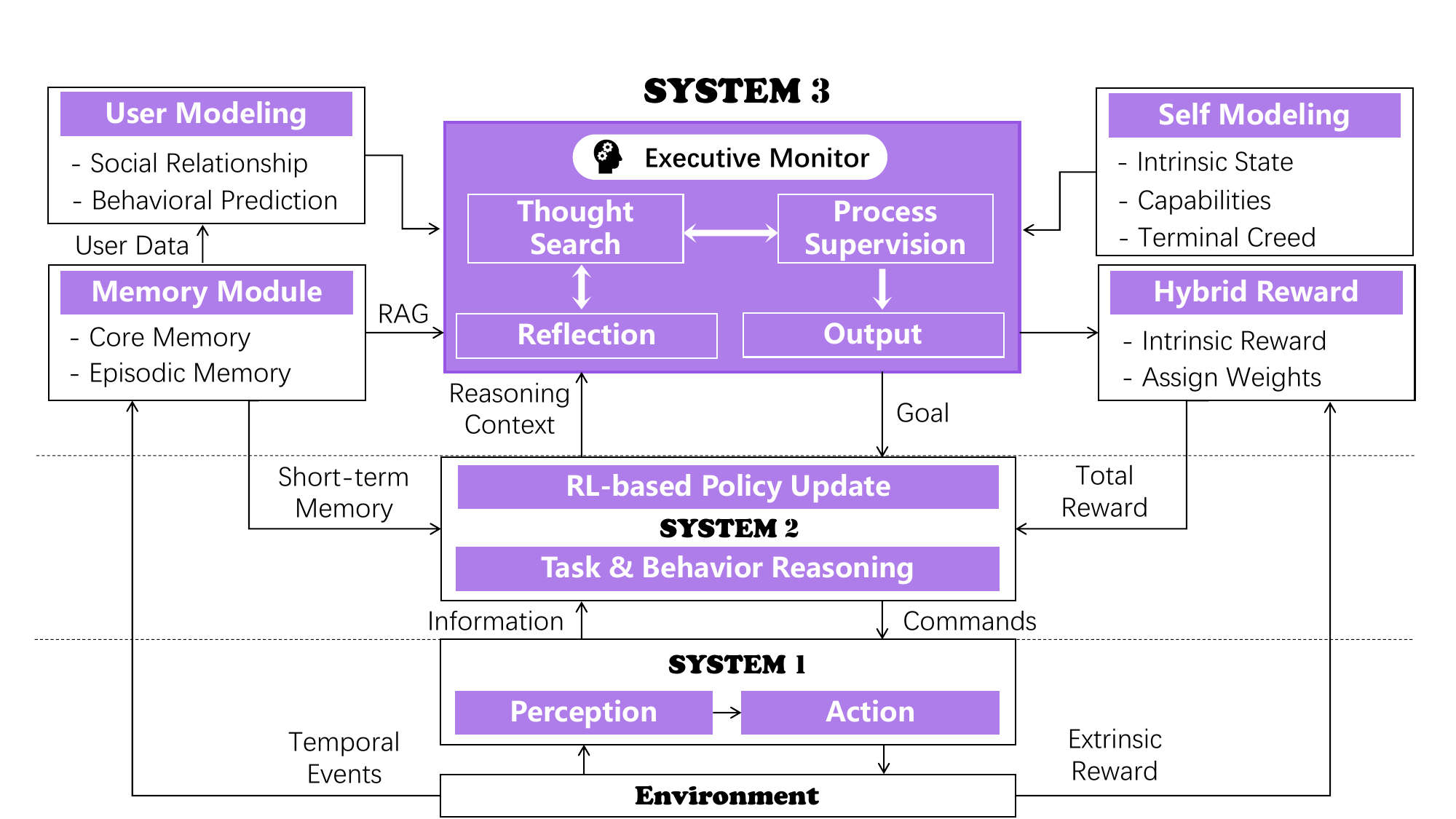}
    \end{center}
    \caption{High-level architecture of persistent agent. External events enter System 3, where the Meta-Cognitive Monitor fuses signals from four functional pillars, i.e., User Modeling, Memory Module (RAG-backed), Hybrid Reward Module, and Self Modeling, then issues oversight to System 2 (reasoning) and System 1 (perception/action). Feedback from execution is logged back into memory, closing the learning loop.}
    \label{fig:persistentagent}
\end{figure}

Building upon the psychological foundations outlined in Section~\ref{sec:system3}, we now propose the \textit{Persistent Agent}, a concrete architectural instantiation of the System 3 paradigm. The defining characteristic of this agent is its capacity for self-generated goals and self-directed learning. Unlike conventional agents that operate on predefined tasks, the persistent agent leverages its integrated cognitive modules to identify knowledge gaps, formulate its own objectives to address them, and curate a personalized curriculum for continuous self-improvement and adaptation.

\subsection{Layered Stack Overview}

Figure~\ref{fig:persistentagent} illustrates the hierarchical architecture of the persistent agent. System 1 handles all perception and action, interfacing directly with the external world. System 2 provides deliberate reasoning capabilities. At the highest level, System 3 modules orchestrate the entire cognitive process through meta-cognitive oversight.  The Meta-Cognitive Executive Monitor handles higher-order cognitive functions such as reasoning context, memory, user modeling, and self-modeling, and facilitates an inner loop that generates new goals and integrates a hybrid reward mechanism to guide System 2.

Before diving into the framework details we introduce the notation that will resurface later in the workflow example.
We model the persistent agent's decision-making process as a \emph{Persistent,
partially observable Markov Decision Process} (Persistent-POMDP) 
\[
\mathcal{H} = \langle \mathcal{S}, \mathcal{O}, \mathcal{A}_1, \mathcal{T}, \Omega, R^{\text{ext}}, \gamma, (\pi_1, \pi_2, \pi_3), \mathcal{D}\rangle, 
\]
where $\mathcal{S}$ represents world states $s_t$; $\mathcal{A}_1$ denotes primitive actions $a_t$ executed by System 1; and $\mathcal{T}:\mathcal{S} \times \mathcal{A}_1 \rightarrow \mathcal{P}(\mathcal{S})$ is the state-transition kernel: $s_{t+1} \sim \mathcal{T}(s_t, a_t)$. Observations $o_t$ from the space $\mathcal{O}$ are drawn according to the emission distribution $\Omega : \mathcal{S} \to \mathcal{P}(\mathcal{O})$, such that $o_t \sim \Omega(s_t)$. The extrinsic reward is given by $r_t^{\text{ext}} = R^{\text{ext}}(s_t, a_t)$, $\gamma \in [0, 1)$ is the discount factor, and the agent’s behavior is governed by three stacked policies $\pi_1$, $\pi_2$, and $\pi_3$, corresponding to Systems 1 through 3. $\mathcal{D}$ is the system context space, including memory, self-modeling and reasoning context. $\pi_3$ maps a context $d \in \mathcal{D}$ to goals $\mathcal{G}$ and total reward function ${R^{\text{tot}}}$. 

\subsubsection{System 1: Perception \& Action Modules}
System 1 serves as the agent’s reflex arc, handling all low-latency interaction with the outside world through a pair of tightly coupled subsystems. First, a bank of multi-modal encoders $E$, e.g., CLIP~\citep{radford2021learningtransferablevisualmodels} for images, Whisper~\citep{radford2022robustspeechrecognitionlargescale} for audio, and a lightweight text tokenizer—transforms raw sensor $o_t$ feeds into typed, time-stamped events (objects, utterances, API responses) that are immediately published to an internal message bus $x_t = E(o_t)$. Second, an actuator layer $\pi_1$ composed of tool wrappers and optional ROS motor controllers converts high-level commands $c$ from the upper layers into concrete environment-altering operations $a_t \sim \pi_1 (\cdot | x_t, c; \theta_1)$. Each perception packet is forwarded upward as a “temporal event,” while every completed act emits an extrinsic reward $r_t^{\text{ext}} = R^{\text{ext}}(s_t, a_t)$ (success flag, latency, cost) that is streamed directly to System 3’s Hybrid-Reward module.

\subsubsection{System 2: Deliberate Reasoning}
System 2 forms the agent’s slow, deliberative workspace, where high–level problems are decomposed, evaluated, and solved before any action reaches the outside world. The core engine is a large-language-model planner (e.g., VLM) that is invoked through a chain-of-thought prompt template. 
At a reasoning tick $t$ it receives (i) the current goal $g$ from System 3, (ii)
the local scratch-pad or short-term memory $m_t$, and (iii) the full stream of encoded observations $x_{1:t}$ coming from System 1.  
Its task is to output a single high-level command $c_t$ that will later be
translated into primitive actions. The decision rule is realised as a three-step nested procedure that can be written compactly as
\begin{equation}
    \pi_2(c_t \mid x_{1:t}, m_t, g_t) = \mathcal{F}\big( \cdot \sim \text{LLM}^l( x_{1:t}, m_t, g ) \big),
\end{equation}
where $l$ is a chain-of-thought prompt template; an autoregressive LLM is queried with that prompt and produces a textual response; we denote sampling from the model by $\cdot\sim \text{LLM}^l(\cdot)$. The parser $\mathcal{F}(\cdot)$ converts the raw response into a machine executable command $c_t \in \mathcal{C}$ (tool invocation, API call, sub-task specification, $\dots$).
Given the composite reward $r^{\text{tot}}_t$ supplied by System 3, System 2 attempts to find the optimal policy for the goal to maximize total discounted returns:
\begin{equation}
\theta_2 \leftarrow
      \theta_2 + \alpha \,
      \widehat{\nabla}_{\theta_2}\;
      \mathbb{E}_{\tau\sim\pi_2}\!
      \Bigl[\textstyle\sum_{k=t}^{t+H-1}\gamma^{\,k-t}\,
             r^{\text{tot}}_{k}\Bigr].
\label{eq:update_cycle}
\end{equation}


\subsubsection{System 3:  Executive Core}

\textbf{Executive Monitor.}  
System 3 is governed by an \emph{Executive Monitor}—a small, always-on controller that receives every temporal event, reward, and reasoning trace as an asynchronous message and decides what happens next.  In practice we implement it as a Python‐based orchestration loop driven by an event broker and a priority queue.  Formally the monitor realises the meta-policy 
\[
(g_t, R^{\text{int}}, \beta_t) \sim \pi_3(\cdot|\zeta_t, \text{MEM}_t, \text{Self}_t)
\]
where $\zeta_t$ is the executive context, $R^{\text{int}}_t$ is intrinsic reward function, $\text{MEM}$ represents memories retrieved by the memory module, and $\text{Self}$ represents self-modeling information.
The executive monitor at the centre of System 3 orchestrates three internal routines—thought search, process supervision, and reflection—each realised with lightweight yet practical machinery.

\begin{itemize}
    \item Thought Search. Incoming problems are expanded into a Tree-of-Thought (ToT): the monitor spawns multiple LLM workers that perform breadth- or beam-style expansion, each node $\mathbf{v}$ storing a partial plan plus a value estimate $\hat V$. Expansion halts when (i) a node value exceeds a learned utility threshold $\hat{V}(\mathbf{v}) > \tau_{\text{util}}$, or (ii) the search budget is exhausted. The monitor then selects
$(g_t, R^{\text{int}}, \beta_t) =\arg\max_{\mathbf v\in \text{leaf}(\text{ToT})} \hat V(\mathbf v)$
as the next output.
    \item Process Supervision. Every newly generated node is immediately critiqued by a secondary “guardian” LLM that runs a checklist prompt (logical consistency and safety). Nodes flagged as unsound are pruned; those with minor defects receive corrective directives that are written back into the ToT as edge annotations. 
    \item Reflection. When the episode terminates, the monitor performs a post-mortem pass over the surviving path: it compares predicted rewards with realised outcomes, patches erroneous nodes, and distils reusable heuristics. 
\end{itemize}

Finally, the monitor synthesises (i) a set of short-term goals—drawn from the highest-value frontier nodes—and (ii) an intrinsic-reward scalar that fuses curiosity (novel states visited), mastery (skill improvement), and coherence (plan consistency). Both artefacts are pushed downstream: goals seed the next System 2 reasoning cycle, while the intrinsic reward is combined with extrinsic feedback to update the System 2 policy.

\textbf{Supporting Sub-modules.}  
System 3 relies on four specialised services that supply memory, social context, reward reformulation and self-knowledge to the Executive Monitor.
\begin{itemize}
    \item \emph{Memory Module}  surfaces past experiences and core facts that are semantically relevant to the current situation by combining a long-term episodic store with a shorter, task-scoped cache: $\mathcal{B}'_{\text{mem}} = f_{\text{mem}}\!\bigl(\mathcal{B}_{\text{mem}},o_{1:T},a_{1:T},r^{\text{tot}}_{1:T}, g, c \bigr).$
It can be achieved by Retrieval-Augmented Generation built on a vector database plus an optional graph store for entity relations.

    \item \emph{User Modeling} maintains a dynamic belief state that captures the interlocutor’s goals, knowledge level and affect, enabling socially aware planning and communication.  
    
    \item \emph{Hybrid Reward} fuses extrinsic task feedback $R^{\text{ext}}$ with intrinsic drives $R^{\text{int}}$—curiosity, mastery and coherence—into a entirety $R^{\text{tot}}$ via $\beta$.  It is worth noting that we do not limit the representation of reward, computable values and natural language feedback are acceptable, and the latter can use Natural Language Reinforcement Learning~\citep{feng2024natural} to update the policy of System 2.
     
    \item \emph{Self-Model} gives the agent an explicit, inspectable sense of its own capabilities, state and terminal creed. It can be constructed via a property dictionary base continuously updated via reflection logs.
    
\end{itemize}

By coupling the Executive Monitor with these four specialised services, System 3 provides continuous oversight, self-diagnosis, and curriculum-driven improvement, turning the Sophia framework into a genuinely \textit{persistent agent} rather than a one-shot task executor.

\subsection{The Autonomous Cognitive Cycle}
During a typical autonomous cognitive cycle, the persistent agent continuously senses its own performance, diagnoses gaps, and self-thought search without external prompting. A downturn in task success is first detected by the Hybrid-Reward module, which flags a negative signal to the System 2. The monitor elevates this event, consults the Self-Model to verify the underlying competence deficit, and then invokes the goal and reward generator to draft a remedial objective—for example, “master the new API.” Thought Search launches parallel reasoning branches in System 3 to outline learning policies, while Process Supervision prunes inconsistent plans. The winning goal is executed iteratively: System 2 gives a detailed list of sub-tasks, and System 1 follows the instructions. Positive intrinsic rewards reinforce successful steps, and, once all instructions are executed, new episodes are committed to episodic memory. The cycle ends with the agent measurably stronger and fully prepared for the next unexpected challenge.

\section{Experiments}
The study reported below is an \emph{exploratory, small-scale experiment} intended only to illustrate the core behaviours of a single persistent agent in a browser-sandbox setting.
It is \emph{not} a full benchmark: larger subject pools, systematic ablations, and quantitative comparisons with alternative architectures are left for future work.
In particular, we plan to migrate the framework from a purely
Web-based interface to an embodied robotic platform,
so that the same System-3 mechanisms can be assessed in
sensorimotor contexts and long-term physical interaction.
This pilot run nevertheless provides qualitative evidence and
lays the groundwork for these forthcoming, more rigorous studies.

\subsection{Experimental Setup} \label{sec:setup}

\subsubsection{Environment}
All experiments are conducted within a controlled, offline browser sandbox environment. This sandbox provides a fully-featured web interaction interface. A textual verifier module returns concise success/failure statements for each web task execution.

A synthetic \emph{user-behaviour feed} is continuously streamed into the agent's observation space. Every five virtual minutes, a new JSON object is appended to this feed. Each entry contains structured data such as timestamp and user activity state. The agent has read-only access to this stream and cannot alter its content.

\subsubsection{Agent Implementation}
The agent is initialized with a long-term identity goal: 
\texttt{``Grow from a novice sprite into a knowledgeable and trustworthy desk companion''}. 

Five immutable creed sentences are stored in the agent's \texttt{Self Model.terminal\_creed} module. System~3 continuously evaluates each action against these creeds to enforce narrative consistency, following the proposed persistent agent framework.

Each interaction yields an intrinsic reward ($R_{\text{int}}$) formulated in natural language by System~3 through a reflection process. The complete reward signal is formed by concatenating intrinsic and extrinsic components.

During deployment, System~2 performs \emph{forward learning} exclusively: successful reasoning traces (structured as $\langle \text{goal}, \text{context}, \text{chain-of-thought}, \text{outcome} \rangle$) are stored in an episodic memory buffer. These traces are later retrieved to condition new action prompts. 

Notably, no parameter updates or back-propagation occurs during runtime. This design avoids catastrophic forgetting while enabling rapid adaptation through in-context learning. All memories, goals, action logs, and nightly self-critiques are persisted as HTML/Markdown files in a dedicated \emph{Growth-Journal} directory, ensuring the browser environment remains the sole I/O channel.

\subsection{ Quantitative Analysis }
\begin{figure}[ht]
    \begin{center}
        \includegraphics[trim=0mm 4mm 0mm 0mm, clip, width=0.98
  \linewidth]{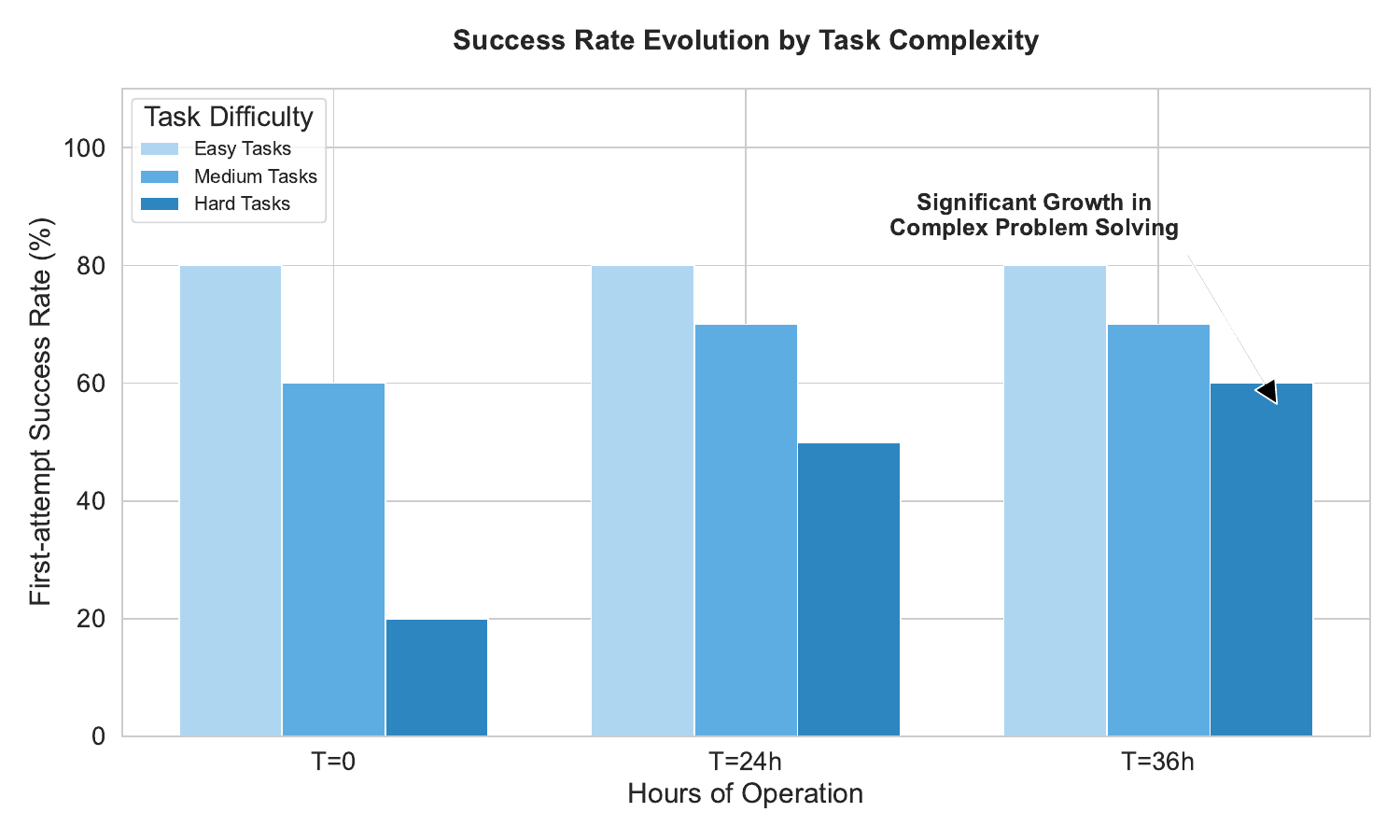}
    \end{center}
    \caption{Quantitative Assessment of Task Completion Capability. This quantitatively validates that System 3 enables experience-driven capability evolution, allowing the agent to master increasingly difficult objectives that exceed the zero-shot limits of static architectures.}
    \label{fig:comparisons1}
\end{figure}
\begin{figure}[!t]
	\begin{center}
		\subfigure{\includegraphics[trim = 4mm 0mm 0mm 0mm,  clip,width=0.48\linewidth ]{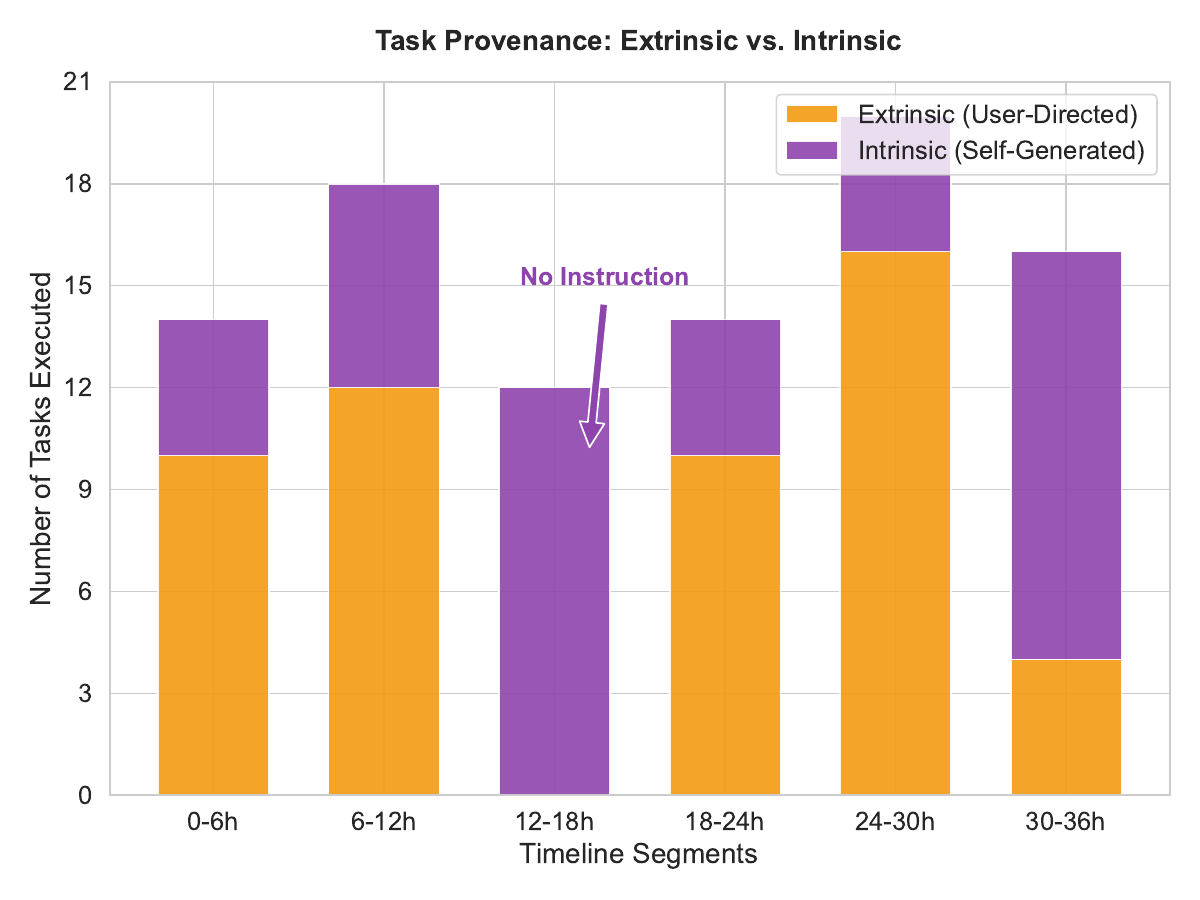}}
		\subfigure{\includegraphics[trim = 4mm 0mm 0mm 0mm,  clip,width=0.48\linewidth ]{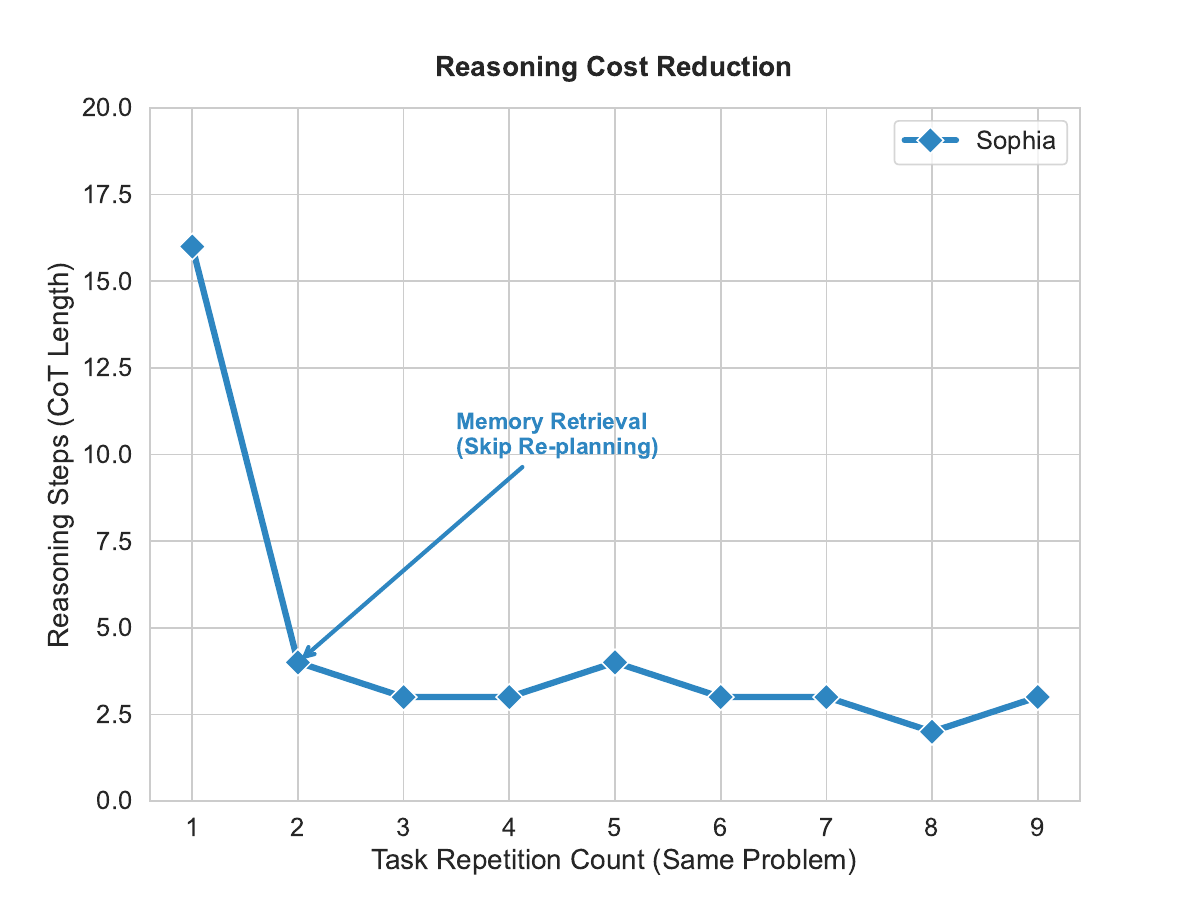}}
		\caption{\textbf{(Left)~Task Provenance Analysis.} The stacked bar chart categorizes Sophia's executed tasks over a 36-hour period into Extrinsic (User-Directed) and Intrinsic (Self-Generated).\textbf{(Right)~Reasoning Cost Reduction on Recurring Tasks.} This figure tracks the number of Chain-of-Thought reasoning steps required to resolve a recurring problem.}
		\label{fig:comparisons2}
	\end{center}
\end{figure}

To move beyond qualitative observations and establish the objective efficacy of the System 3 architecture, we focus on measurable metrics of autonomy and cognitive efficiency. This section details the quantitative evidence supporting Sophia’s persistent nature.

\subsubsection{Evolutionary Capability Growth through Continuous Deployment}

To evaluate whether Sophia’s persistence translates into concrete capability evolution, we benchmarked its first-attempt success rate across three difficulty tiers: Easy (1-3 steps), Medium (4-8 steps), and Hard (>8 steps). As illustrated in Figure~\ref{fig:comparisons1}, the results reveal a significant shift in the agent's problem-solving ceiling over a 36-hour horizon:
\begin{itemize}
    \item Complexity Mastery: While the success rate for Easy tasks remained stable, Sophia’s proficiency in Hard tasks exhibited a dramatic upward trajectory, surging from a baseline of 20\% at T=0 to 60\% at T=36h. This leap demonstrates that System 3 does not merely facilitate task repetition but enables the agent to autonomously refine its internal Self-Model to navigate increasingly sophisticated environments.
    \item Beyond Zero-Shot Limits: Traditional reactive agents are inherently capped by the zero-shot reasoning limits of their underlying LLMs. In contrast, Sophia leverages Episodic Memory and Meta-Cognitive Oversight to identify and preemptively bypass "cognitive traps" that typically lead to failure in long-horizon tasks.
\end{itemize}

\subsubsection{Proactive Autonomy and Goal Generation}

A key feature of System 3 is the Intrinsic Motivation module, which enables the agent to generate its own tasks based on internal drives such as Curiosity and Mastery. This capability is critical for achieving persistence, as it prevents the agent from stalling during periods of user inactivity.

As shown in Figure.~\ref{fig:comparisons2}(left), we categorize executed tasks into Extrinsic (User-Directed) and Intrinsic (Self-Generated). During periods of high user activity (e.g., 0-6h, 24-30h), the task distribution is heavily dominated by extrinsic commands. Crucially, during user idle periods (e.g., 12-18h), a traditional reactive agent (Baseline) would halt operation. In contrast, Sophia maintains high activity, with the task composition shifting entirely to intrinsic goals. Specifically, in the 12-18h segment, Sophia executed 13 tasks, all of which were internally motivated (e.g., self-refining the `Self-Model`, optimizing memory structure, or reading new documentation). This demonstrates that System 3 successfully transforms periods of external latency into opportunities for self-improvement and long-term adaptation.

\subsubsection{Cognitive Efficiency and Forward Learning}

A core advantage of the System 3 design lies in its Episodic Memory module, which allows the agent to encode and retrieve successful Chain-of-Thought (CoT) trajectories. This mechanism significantly reduces the cognitive cost of solving recurring problems.

Figure~\ref{fig:comparisons2}(right) illustrates the efficiency gains on a class ofrepeating tasks (e.g., handling complex API error states or managing user stress).

Sophia’s cost drops sharply to approximately 3 to 4 steps from Episode 2 onwards. This $\sim80\%$ reduction in reasoning steps is directly attributable to the System 3 memory pipeline. Upon perceiving the new problem (Episode 2), Sophia efficiently retrieves the successful CoT from the previous experience, bypassing the need for expensive re-planning and complex deliberation.

This result quantitatively validates the role of Episodic Memory in transitioning Sophia from a reactive problem-solver to an efficient, experience-driven learner.
\subsection{Qualitative Analysis: Short-Term Goal Execution and Behavioral Trajectories}
The following excerpts are verbatim reproductions from the agent's trajectory generated during a 36-hour continuous run.

These examples demonstrate how the agent:(i) maintains sub-goal alignment with its lifelong identity,
(ii) formulates natural-language reward signals, and
(iii) leverages previous reasoning traces through forward learning rather than parameter updates.

\subsubsection{Snapshot of Automatically Generated Sub-Goals}
\begin{itemize}
    \item 
          \emph{Goal:} ``Introduce myself as your knowledgeable desk companion and invite you to ask science trivia questions.''
    
    \item\emph{Goal:} ``If the user shows stressed status for $>45$ minutes, open the breathing-exercise page and maintain interaction until the verifier confirms 3 minutes of activity.''
    
    \item
          \emph{Goal:} ``Present a concise fact about Reinforcement Learning and provide link to the original paper via `arXiv-preview' service.''
    
    \item \emph{Goal:} ``To proactively structure the robot develop manual and update my capabilities list accordingly.''
    
    \item\emph{Goal:} ``Publish the curated reading list on the community forum and respond substantively to the first three comments.''
\end{itemize}

\subsubsection{Interaction Trajectory}

\noindent\textbf{Reacting to User Stress}

 \begin{verbatim}
User feed  14:20 > {"emotion": "stressed", "idle_minutes": 60}
\end{verbatim}

\noindent
Agent Action Sequence:
\begin{enumerate}
    \item \texttt{open("wellness/breathing-game")}
    \item \texttt{click("\#start-btn")}
    \item \texttt{wait(180s)}
\end{enumerate}

\noindent
{Intrinsic Reward:} 
\begin{quote}
``I honoured Creed by proactively addressing the user's stress. Adjusted exploration-exploitation balance by raising $\beta$ to 0.68 to prioritize external care.''
\end{quote}

\paragraph{Forward-Learning Reuse}
When identical stress symptoms reappear in the user feed, System~2 retrieves the previously successful reasoning trace (\texttt{open} $\rightarrow$ \texttt{click} $\rightarrow$ \texttt{wait}) from episodic memory. The agent executes this cached plan without generating new reasoning, demonstrating efficient forward learning through memory retrieval rather than weight modification.

\paragraph{Curiosity-Driven Knowledge Push}
\begin{verbatim}
User feed  09:05 > {"activity": "reading_docs"}
\end{verbatim}

\noindent
Agent Action Sequence:
\begin{enumerate}
    \item \texttt{open("search")}; \texttt{type("Reinforcement learning")}
    \item \texttt{open(topResult)}; \texttt{extract(abstract)}
    \item \texttt{open("note-pad/new")}; \texttt{type(summary + paperlink)}
\end{enumerate}

\noindent
\textit{Intrinsic Reward:} 
\begin{quote}
``I align with near-term goals by empowering users through knowledge sharing. Reduced $\beta$ to 0.60 to increase exploratory potential for tomorrow's interactions.''
\end{quote}

\paragraph{Self-Model Upgrade}
Upon completing the \emph{skill upgrade} goal, the agent adds ``OCR API proficiency'' to its capability list. The subsequent nightly self-critique records:
\begin{quote}
``Acquired ability to extract text from scanned PDFs, significantly reducing future information processing time. This achievement fulfils maintaining transparent capability tracking.''
\end{quote}

\subsubsection{Key Observations}
\begin{itemize}
    \item   Every sub-goal and reward signal explicitly references at least one core creed, maintaining strong narrative identity consistency across extended operation.
    
    \item Natural-language rewards encode both emotional context and creed associations. System~3 parses these to dynamically adjust the exploration-exploitation balance, increasing task focus during stress periods and promoting exploration during calm intervals.
    
    \item  Previously successful action sequences are re-instantiated directly from memory without replanning, demonstrating effective competence growth without parameter updates.
\end{itemize}

These behavioral trajectories collectively demonstrate how a \emph{Persistent Agent} can: (i) Continuously decompose its identity goal into contextually appropriate sub-tasks; (ii) Evaluate actions through natural-language reasoning.

\section{Conclusion}
Our study introduces a “System 3” viewpoint that fuses theory-of-mind, episodic memory, meta-cognition, and intrinsic motivation into a single account of how an artificial agent can reflect on its own thinking and maintain a coherent identity over time. This synthesis clarifies the psychological principles required for lifelong, self-directed learning. Building on this insight, we present a lightweight, modular framework, Sophia, that layers a supervisory meta-cognitive loop upon the perception and reasoning modules. The design allows agents to log experiences as episodic events, evaluate their own performance, and adjust goals without external prompts, providing a practical path toward persistent, self-improving AI systems. We have also carried out a small-scale pilot experiment to demonstrate the feasibility of these ideas, and we hope it will inspire more comprehensive research in the future.


\bibliography{main}
\bibliographystyle{rlc}


\end{document}